\documentclass{article}
\usepackage{ijcai24}

\usepackage{times}
\usepackage{soul}
\usepackage{url}
\usepackage[hidelinks]{hyperref}
\usepackage[utf8]{inputenc}
\usepackage[small]{caption}
\usepackage{graphicx}
\usepackage{amsmath}
\usepackage{amsthm}
\usepackage{booktabs}
\usepackage{algorithm}
\usepackage[switch]{lineno}
\usepackage{color}
\usepackage{xcolor}
\usepackage[most]{tcolorbox}
\usepackage{multirow}
\usepackage{makecell}
\usepackage{diagbox} 
\usepackage{algpseudocode}
\usepackage{listings}
\usepackage{hyperref}
\definecolor{Emerald}{RGB}{80,200,120} 

\lstdefinelanguage{json}{
    basicstyle=\tt\small,
    showstringspaces=false,
    breaklines=true,
    tabsize=1,
    morestring=[b]{"},
}

\title{Growing from Exploration: A self-exploring framework for robots based on foundation models}

\author{
	Shoujie Li $^1$ $^*$,
	Ran Yu  $^1$ $^*$,
        Tong Wu$^1$ $^*$,
        JunWen Zhong$^2$,
        Xiao-Ping Zhang$^1$ $^3$,
        Wenbo Ding$^1$ $^4$ $^\dagger$,
 }

\author{
Shoujie Li $^1$ $^*$,
Ran Yu  $^1$ $^*$,
Tong Wu$^1$ $^*$,
JunWen Zhong$^2$,
Xiao-Ping Zhang$^1$ $^3$,
Wenbo Ding$^1$ $^4$ $^\dagger$,
\affiliations
$ ^1$ Tsinghua Shenzhen International Graduate School, Tsinghua University, Shenzhen 518055, China.\\
$^2$ Department of Physics and Chemistry, Faculty of Science and Technology, University of Macau, Avenida da Universidade, Taipa, Macau 999078, China.\\
$^3$ Department of Electrical, Computer, and Biomedical Engineering, Ryerson University, Toronto  ON M5B 2K3, Canada.\\
$^4$ RISC-V International Open Source Laboratory, Tsinghua-Berkeley Shenzhen Institute, Shenzhen 518055, China.\\
$*$ These authors contributed equally to this work.\\
$\dagger$Corresponding authors: Wenbo Ding.
\emails
\{lsj20, yur23\}@mails.tsinghua.edu.cn,
sympathischmann35@gmail.com,
junwenzhong@um.edu.mo,
xzhang@ee.ryerson.ca,
ding.wenbo@sz.tsinghua.edu.cn,
}

\begin{document}

\maketitle

\begin{abstract}

Intelligent robot is the ultimate goal in the robotics field. Existing works leverage learning-based or optimization-based methods to accomplish human-defined tasks. However, the challenge of enabling robots to explore various environments autonomously remains unresolved. In this work, we propose a framework named GExp, which enables robots to explore and learn autonomously without human intervention. To achieve this goal, we devise modules including self-exploration, knowledge-base-building, and close-loop feedback based on foundation models. Inspired by the way that infants interact with the world, GExp encourages robots to understand and explore the environment with a series of self-generated tasks. During the process of exploration, the robot will acquire skills from beneficial experiences that are useful in the future. GExp provides robots with the ability to solve complex tasks through self-exploration. GExp work is independent of prior interactive knowledge and human intervention, allowing it to adapt directly to different scenarios, unlike previous studies that provided in-context examples as few-shot learning. In addition, we propose a workflow of deploying the real-world robot system with self-learned skills as an embodied assistant. Project website: \href{https://sites.google.com/view/gexp}{\textcolor{blue}{GExp.com}}.

\end{abstract}

\section{Introduction}

\begin{figure}[t]
\centering
\includegraphics[width=0.5\textwidth]{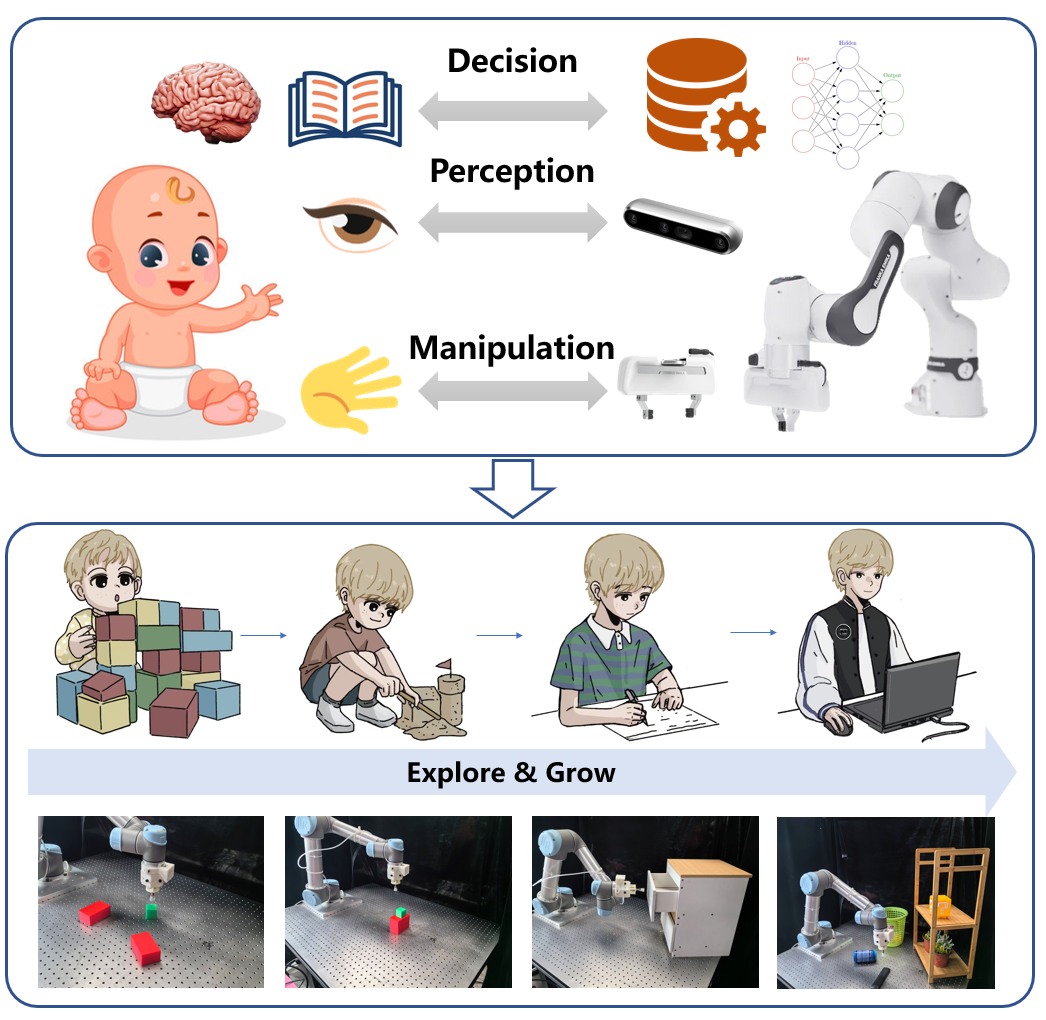}
\caption{\textbf{An infant-inspired framework for autonomous robotic exploration (From left to right: pick blocks, stack blocks, open drawer, classify objects). }}
\label{fig:1}
\end{figure}

Not only for humans\cite{hacques2021exploring,ivaldi2013object}, autonomous exploration and learning are equally important for robots\cite{balsells2023autonomous}, which is a sign of robot intelligence\cite{wahde2012introduction}. Autonomous robots can not only liberate people from hard or dangerous labor but also complete some daily and ubiquitous tasks, such as searching, cooking, and cleaning. However, the design of robots with autonomy is very challenging. While robots can complete some tasks with the help of reinforcement learning\cite{deng2019deep}, imitation learning\cite{hussein2017imitation}, and other methods, they are only capable of dealing with related tasks and cannot learn by themselves.  Despite these methods have achieved outstanding results, they constrain the automatic learning ability of robots and it is difficult to achieve real intelligence only with these methods.

Recently, the birth of Large Language Models (LLMs), such as GPT-4 \cite{openai2023gpt} and Llama 2 \cite{touvron2023llama}, provides new paths to intelligence. LLMs have not only propelled the development of artificial intelligence technology but have also significantly enhanced the decision-making and task-planning capabilities of robots. There is a growing body of research utilizing LLMs to address existing robotics challenges, as evidenced in studies including\cite{huang2023voxposer,liang2023code,brohan2023can,wang2023gensim}. However, most existing methods rely heavily on detailed, complete prior information and the provision of human-specific in-context examples. This dependency often limits the robots' generalization capabilities across diverse environments.

On the other hand, there are studies demonstrating the potential of creating autonomous agents based on LLMs, which are capable of independently and continuously resolving problems \cite{zhu2023ghost,wang2023voyager}. A notable example is Voyager \cite{wang2023voyager}, which presents an embodied agent operating within a Minecraft virtual environment. These agents, built upon LLMs, interact with their environment through textual feedback, which hinders practical applications in robotics. The recent advancements in vision-language models have been particularly significant, showcasing the potential for general object recognition and understanding of spatial relationships \cite{hu2023look,yang2023dawn}. This progression opens up new possibilities for robotics, allowing for more nuanced and adaptable interactions with various environments.

To facilitate autonomous exploration and learning in robots, we introduce an innovative framework named GExp, which is predicated on the use of foundational models. As shown in Fig.\ref{fig:1}, the framework is inspired by the growth process of infants.  Newborn babies begin with an incomplete understanding of the world's logic, gradually perceiving the world through their vision and interacting with it physically, leading to progressive development. Similarly, in GExp, pre-trained foundational models are employed to assist the robot in navigating and learning within unfamiliar environments. With an understanding of the environment with VLM, robots will generate tasks and try to solve them with the ability of LLM. Throughout the exploration phase, the robot will learn from successful experiences by generating relevant skills in a zero-shot manner. These skills are continuously collected into a comprehensive library, allowing the robot to tackle increasingly complex scenarios.

The contribution of the article mainly includes:

\begin{itemize}

    \item A framework for robot autonomous self-exploration is proposed, which does not require the imposition of specific human-defined tasks, nor does it necessitate any prior knowledge about the environment. Its primary function is to facilitate continuous and independent exploration by the robot.

    \item In our approach, we enable robots to learn during exploration by leveraging successful experiences. This is achieved through the use of LLM, which assists in generating general skills in a zero-shot manner. The skills thus developed not only enhance the robot's capacity to solve increasingly complex tasks encountered during exploration but also expand the boundaries of its abilities.

    \item We created a self-verification module using a pre-trained VLM to analyze and determine task execution success. This evolves into "backtracking control" enabling the robot to verify each task step's success by assessing preconditions. This ensures actions align with the overall task objective, enhancing the robot's precision and effectiveness.

    \item To evaluate the exploration and self-learning capabilities of the robot effectively, we conduct a series of experiments specifically designed to validate the feasibility and effectiveness of our proposed framework.
\end{itemize}

\section{Related Works}
\textbf{Foundation Models For Robotics.}  The evolution of Large Language Models (LLMs) has significantly transformed the robotics field, particularly in areas such as task planning \cite{vemprala2023chatgpt,huang2022inner}, robot action generation \cite{kwon2023language}, and code development \cite{liang2023code,xu2023creative}. These models have proven effective in integrating with traditional motion planning algorithms \cite{huang2023voxposer} or reinforcement learning \cite{ma2023eureka}, enhancing capabilities in multi-modal information processing, multi-robot coordination, and human-robot collaboration. Another kind of foundation model, vision-language models (VLMs) \cite{kirillov2023segment,minderer2205simple,vemprala2023chatgpt}, also fast developed recently and many researches have been conducted to discuss its impact in robotics \cite{hu2023look}. Our work combines the LLM and VLM to construct an autonomous embodied agent with the ability to explore the real world. \newline

\textbf{Language Models For Task and Motion Planning.} Long-horizon planning problem is one of the key research areas in robotics \cite{garrett2021integrated}. Conventional methods are mainly based on optimization and reinforcement learning methods \cite{nau1999shop}. Recently, some researchers utilized LLM as the core of robots to divide sub-tasks or generate high-level plans for task and motion planning \cite{brohan2023can,huang2022inner,xu2023creative,xie2023translating}. To help the LLM better understand the given task and reduce the hallucinations in generating plans, in-context examples are used in most of the relevant works. However, this could reduce the generalization ability, since those examples should be collected manually for a given environment. We are trying to improve the generalization ability of LLM-based planning problems by designing a framework for robot self-exploration.

\section{Method}
\subsection{Grow from Exploration}
Assuming a robot faces an unfamiliar environment, it is not provided with any human-specific tasks and prior knowledge about the environment. The robot can observe the environment with an RGB-D camera and interact with the objects in the environment with a robot arm. Our goal is to let robots autonomously interact with the environment to explore the boundaries of their abilities. To reach that goal, a robot should understand the environment, generate tasks by itself, make plans, and interact with the environment to find out the feasibility of generated plans.

\begin{figure*}[htbp]
\centering
\includegraphics[width=1\textwidth]{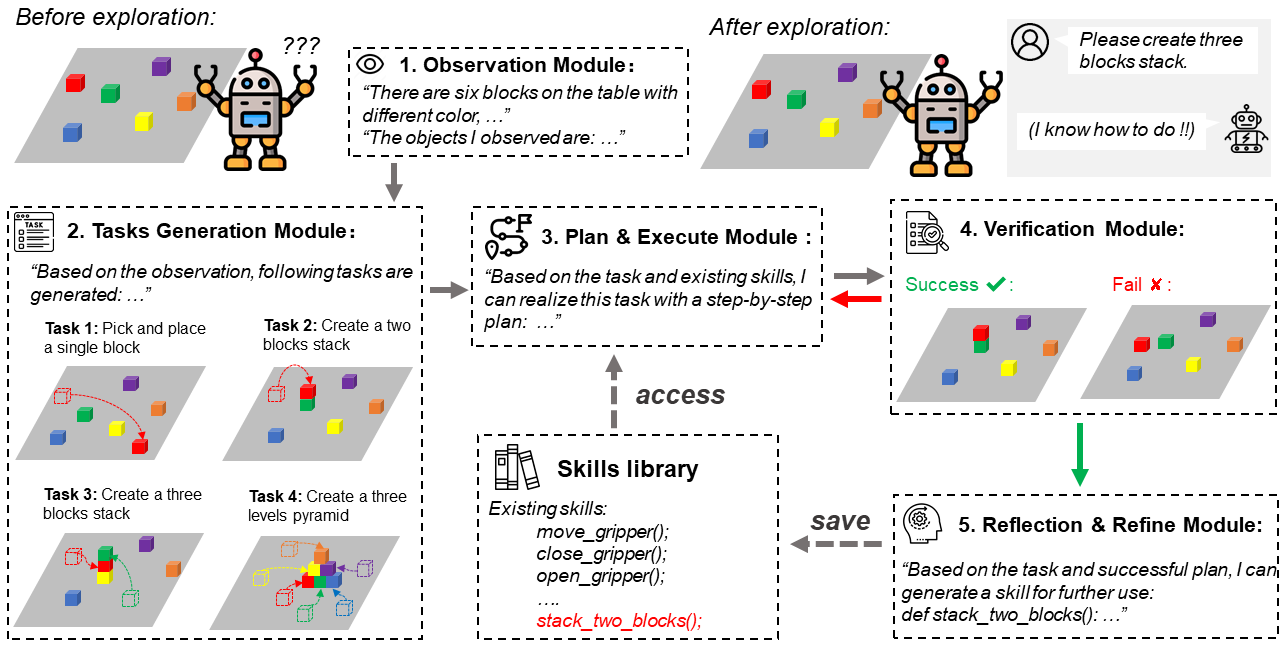}
\label{fig Robot Exploration Framework}
\caption{\textbf{Robot exploration framework GExp.} GExp allows robots to actively explore the new environment by observing the environment, generating feasible tasks, solving those tasks independently, and verifying the success of execution. During the process of exploration, GExp maintains and updates a skills library by reflecting on successful tasks and generating task-related skills. }
\end{figure*}

\subsubsection{\textit{A. Scene Understanding and Tasks Generation}}
In our setting, the robot does not have access to any prior information about the environment. Therefore, robots need to first perceive and understand their surroundings. We leverage VLMs to characterize the information in the scene with an image as input. The purpose of scene understanding can be divided into two parts: identifying observed objects with their attributes and generating relation descriptions that explain the relative positions of different objects. 

After obtaining the scene description and information about the objects in the space, the robot will ask LLM to generate a series of manipulation tasks to explore the environment and possibly feasible skills. A task is characterized by its name, related objects, and a detailed description. To ensure efficient task generation, we employ an object-related principle. This principle stipulates that each generated task must contain at least one observed object, preventing LLM from generating some useless tasks. Inspired by the growth of infants and curriculum learning, we also prompt LLM to generate structured tasks from simple to complex. Simpler tasks typically involve operations on individual objects and are characterized by straightforward descriptions and a limited number of steps. Complex tasks involve multiple objects and normally relate to long-horizon planning.

\subsubsection{\textit{B. Planning and Executing}}

LLMs have strong reasoning and planning abilities. It has also been proved that LLMs are capable of grounding a high-level plan into several sub-actions. Motivated by these, we developed a module that uses LLM as a planner and controller. LLM is given a self-generated task and a library of acquired skills and outputs a high-level plan with an action sequence that is composed of the acquired skills. These skills can be in any form, such as trained policy through Reinforcement Learning Imitation Learning, or just code. In this paper, we choose code as the form of skills for simplicity. To improve the reasoning ability of LLM, we utilize chain-of-thought (CoT) in the process of plan generation as \cite{wei2022chain} do. In this paper, we choose code as the form of skills, as code can be directly generated by LLM without extra training or human demonstrations. However, using code for grounding has limitations in complex motion planning, which is not the main focus of our work.

During the exploration, the robot will also maintain a skill library that contains all acquired skills. The skill library is kept updated by LLM generating a new one when the task is completed. The following functions are predefined in the initial skill library as primitive functions:
\begin{itemize}
    \item \textbf{movep(position):} Move the end-effector of the robot to a given position.
    \item \textbf{close\_gripper():} Close the gripper to grasp an object. The gripper is initially open.
    \item \textbf{open\_gripper():} Open the gripper to place an object.
    \item \textbf{get\_obj\_position(object\_name):} Get the position of an object with the given name. The output of this function is a tuple of object position (x, y, z).
    \item \textbf{go\_home():} Move the gripper to the home position.
\end{itemize}

The choosing principles of those are generalization and simplicity. From the perspective of generalization, those skills should be required by most manipulation tasks and can be easily defined for any robot system.  From the perspective of simplicity, we want to inspire robots to generate more high-level and task-specific skills based on those simple skills, instead of heavily relying on human-specific actions. Except for those functions, we also define the robot arm's available working space for safety reasons:
\begin{itemize}
    \item \textbf{BOUNDS:} The available working space in three dimensions, [[xmin, xmax], [ymin, ymax], [zmin, zmax]].
\end{itemize}

\subsubsection{\textit{C. Verification and Error Correction}} \label{section self-verification}

After implementing the generated plan, it is crucial to verify its success. The GExp operates autonomously based on foundation models, thus excluding any human involvement in confirming the success of the implemented manipulation plan. To ascertain whether the task has been completed successfully, foundation models serve as success detectors.

We utilize two different methods for self-verification:  code-based method and vision-based method. The former one, utilizing the reasoning and code-creating ability of LLM, generates a success detection function based on the description of the given task. The vision-based method leverages advanced VLMs (GPT4-V) to judge whether the task requirement is satisfied. It is worth mentioning that existing VLMs have limited capability in object detecting, semantic grounding, and relation understanding. To improve its performance, we provide VLM with a final state observation image and initial state description (generated at the scene understanding stage). This can significantly help VLM make correct answers in success verification.

In the event of a plan failure due to errors, GExp will attempt to solve them. Errors can be divided into two categories: \textbf{interpretation errors} and \textbf{grounding errors}. Interpretation errors occur during the code generation phase and include syntax errors, which breach programming language rules, and invalid instance errors, characterized by the use of undefined variables or functions. The grounding errors are associated with the execution phase of the code. Common grounding errors include manipulating with object in a gripper and directional misjudgments.

Our approach to addressing these two types of errors differs. For interpretation errors, we solve them by letting LLM regenerate the code without changing the plan. This is achieved by providing the LLM with the error code with a corresponding error message. Results show that in most cases, these interpretation errors can be rectified in a single iteration. The second type of error, grounding error, can be detected by self-verification but hard be corrected by LLM itself without human intervention. We take an iterative plan-exploration strategy: regenerating the high-level plan by modifying the existing one. LLM is required to modify the current plan. However, it should be mentioned that not all self-generated plans are feasible to solve or exceed the current capabilities of foundation models.

\subsubsection{\textit{D. Reflection and Skill Acquirement}}
If the success of the task has been verified, LLM is required to reflect the generated plan and code to refine a skill function with the following rules:
\begin{itemize}
    \item The created function should be general, which makes sure the function can be more likely to be used in further tasks.
    \item The LLM should generate the description of the new function, including its usage, input, and output.
    \item The current task should be resolved by using a new function, which serves as an example for further utilization.
\end{itemize}
Once the function has been generated, it should be first checked with no syntax error and then updated to the skills library.

\subsubsection{\textit{E. Prompting Design}}
The following rules are followed when designing the prompt:
\begin{itemize}
    \item \textbf{Format regularization:} To ensure the stability of the system, all outputs of LLM follow the JSON format. This is realized by providing a format example in the system prompt. 
    \item \textbf{Prompt iteration:} When the robot generates a new skill function from the interaction with the environment, the prompt should iterated to keep consistent with the current skills library.
    \item \textbf{Generalized prompt:} As one of the main motivations, we do not include any prior information about the environment and in-context examples. The framework can be applied to different environments without modifying the prompt.
\end{itemize}

\begin{algorithm}[t]
    \caption{Learn Skills through Exploration}
    \label{alg:algorithm}
    \textbf{Input}: Initial visual observation $x_0$, initial skills library with basic actions $\Pi=\Pi_0$, scene language description $l$, observed objects $O_{1:j}$\\
    \textbf{Output}: Refined skills library $\Pi$
    \begin{algorithmic}[1] 
        \State $l$, $O_{1:j}$ = VLM($x_0$) \Comment{See Appendix \ref{appendix section 1}}
        \State $T_{1:k}$ = LLM($l$, $O_{1:j}$) \Comment{See Appendix \ref{appendix prompt generate tasks}}
        \For{$T_i$ in $T_1, ..., T_k$}
        \State $T = T_i$, $t = 1$
        \While{$t < $ num\_retries}
        \State $p$, $a_{1:N}$ = LLM($T$, $\Pi$) \Comment{See Appendix \ref{appendix prompt planning}}
        \State Execute plan $p$ with actions $a_{1:N}$
        \State flag\_success = VLM($x_i$) \Comment{See Appendix \ref{appendix prompt verify}}
        \If {flag\_success}
        \State $\pi$ = LLM($p$, $a_{1:N}$, $\Pi$) \Comment{See Appendix \ref{appendix prompt learn skill}}
        \State $\Pi = \Pi \cup \{\pi\}$
        \State break
        \Else
        \State t = t + 1
        \EndIf
        \EndWhile
        \EndFor
        \State \textbf{return} Refined skills library $\Pi$
    \end{algorithmic}
\end{algorithm}

\subsection{How Acquiring Skills Strengthen Robot?}
Foundation models have been pre-trained with tons of data from the internet. However, when they are used for planning complex real-world problems, their performances are influenced by hallucinations. The performance can be improved by providing in-context examples by humans, while we want the robot to enhance its ability by learning from experiences in exploration. 

Denote the set of manipulation tasks can be performed by combining provided primary functions is $S^{*}_{\Pi_0} =\{T^{*}_1, T^{*}_2, ..., T^{*}_n\}$, the probability that LLM successfully finish the task $T^{*}_i$ is $P_{LLM}(T^{*}_i|\Pi_0)$. The goal of self-exploration is enabling the robot to automatically learn several skills $\Pi$ during the process of interaction between real-world environments to make $\sum_i{P_{LLM}(T^{*}_i|\Pi)}>\sum_i{P_{LLM}(T^{*}_i|\Pi_0)}$.

As introduced in the previous sections, the skill-collecting module can facilitate the exploration process. The robot can use previously generated skills to solve more complex problems. The generated skills form a tree structure: any new skill can be realized by the combination of basic skills (functions), and generated skills are also interrelated to each other. When the creation of skills is finished after one or multiple explorations, the skills library can be edited manually by deleting redundant skills or further distilled with LLM when combined with other skills libraries. 

\subsection{Deployment after Exploration with Backtracking Control} \label{section deployment}

After the robot explores one or more scenes, it is equipped with a self-learned skills library and able to complete user-given tasks within its capabilities. Different from exploring with self-generated tasks, the robot is provided with free-from-language instruction, such as "Clean the table" and "Where is my cup".

To deploy the robot after exploration, we utilize the workflow shown in Figure \ref{figure deployment}. Inspired by the idea of ReAct \cite{yao2022react}, the robot is controlled by a controller powered by LLM which will continuously switch between \textbf{observer module} and \textbf{executor module}, until satisfying the user's requirement. When receiving user instructions, the controller will judge if the user has been provided with enough information to generate an executable plan. When information is sufficient, the controller will call the executor module with an executable task as input (same format as the self-generated one in exploration). The executor module will first analyze the given task referring to the whole skills library, and then generate a high-level plan with controlling code. Otherwise, the observer module will be evoked to ground the user specification with environment information. The input of the observer module is a query string generated by and the observer will try to answer it based on VLM. This process will continue until the controller believes the user requirement is satisfied or can not be realized.

When a robot executes its generated plan, there exists the possibility that some steps are not successfully implemented due to errors and online disturbances. We propose backtracking control in the executor module, which automatically generates a precondition for each step, verifies the result of each step, and backtracks to the appropriate step. As shown in Fig, LLM will generate a precondition for each step by considering its context such as the target task and plans for completing this task. We organize the preconditions as object-description text, characterizing the relation between objects or the state of objects. For example, "plate on the microwave" describes the relationship between plate and microwave, and "top drawer is open" describes the state of the top drawer. The verification process utilizes VLM, which is the same as vision-based task verification introduced in Section \ref{section self-verification}. If the precondition is not satisfied, the robot will retry by backtracking, that is going back to the nearest step whose precondition meets the current state. Then robot will start from that step and retry the task. We find that the successful rate and robustness of completing a task are significantly improved by absorbing such a close-loop method.

\begin{figure}[ht]
\centering
\includegraphics[width=0.5\textwidth]{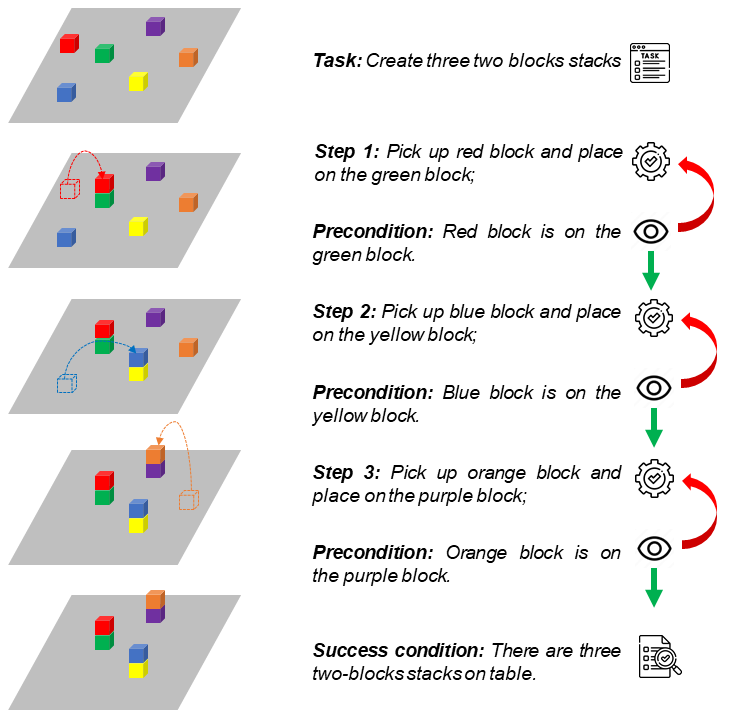}
\caption{\textbf{The proposed backtracking control method.} }
\label{figure backward control}
\end{figure}

\section{Experiments}
The experiments are designed to validate the following questions:
\begin{itemize}
    \item Is the proposed framework capable of efficiently guiding a robot to proactively engage with its surroundings and continually develop new skills, thereby enhancing the exploration process?
    \item Whether the proposed framework has a certain degree of generalization ability? This means that it can be applied to different scenarios and collect useful skills. These skills should be further used to solve related tasks after exploration.
    \item Can we deploy a robot system in the real world that utilizes skills learned from GExp to accomplish a task specified by the user?
\end{itemize}

\subsection{"BLOCKS WOLD"}
For the first question, we build the "BLOCKS WOLD" table-top simulation scenario based on RAVENS \cite{zeng2020ravens}, where six blocks with different colors are placed on the table and a UR5 robot arm is used to interact with the environment. We use "BLOCKS WOLD" for the experiment 
An ablation experiment is conducted during the process of self-exploration. We compare the proposed method GExp with a framework without collecting skills and a framework without self-verification.

\begin{table}[h]
\centering
\begin{tabular}{m{3.3cm}<{\raggedright}m{1.4cm}<{\centering}m{1.5cm}<{\centering}m{0.7cm}<{\centering}}
\hline
\textbf{Task} &  \textbf{w/o learn skills} &  \textbf{w/o self-verification} & \textbf{GExp}\\
\hline
Move one block & \textbf{0.8} & 0.4 & \textbf{0.8} \\
One two-blocks stack & 0.6 & 0.9 & \textbf{1.0}\\
Three two-blocks stack & 0.4 & \textbf{1.0} & \textbf{1.0}\\
One three-blocks stack & 0.1 & 0.7 & \textbf{0.8}\\
Three blocks pyramid & 0.0 & 0.5 & \textbf{0.7}\\
Six blocks pyramid & 0.0 & 0.4 & \textbf{0.7} \\
Average & 0.32 & 0.25 & \textbf{0.83} \\
\hline
\end{tabular}
\caption{\textbf{Ablation evaluation results of GExp in "BLOCKS WOLD".} The value in the table represents the successful rate of 10 runs. Results show that skills learning and self-verification can facilitate the process of exploration.}
\label{table RAVENS ablation results}
\end{table}

Six different tasks are generated by GExp in "BLOCKS WOLD": pick-and-place single block, one two-block stack, one three-block stack, three two-block stacks, a pyramid with three blocks, and a complex pyramid with six blocks. The details of generated tasks can be found in Appendix \ref{appendix tasks in blocks world}. After accomplishing one task, GExp will generate a related skill for further use (see Appendix \ref{appendix skills in blocks world}).

We report the results in Table \ref{table RAVENS ablation results}. As the beginning task, moving one block could not be realized successfully because the robot moved the block before lifting it. The self-verification module could help to correct the plan by regenerating it. The following three tasks (blocks stack) show the significant difference between planning with and without using self-generated skills. The last two tasks (pyramid building) go beyond the planning and reasoning ability of LLM with predefined skills. However, GExp enables the robot to solve these complex, long-horizon planning challenges. The average successful rate shows that GExp achieves 0.51 and 0.58 improvement compared with removing skills learning and self-verification modules respectively. 

\begin{figure*}[h]
\centering
\includegraphics[width=1\textwidth]{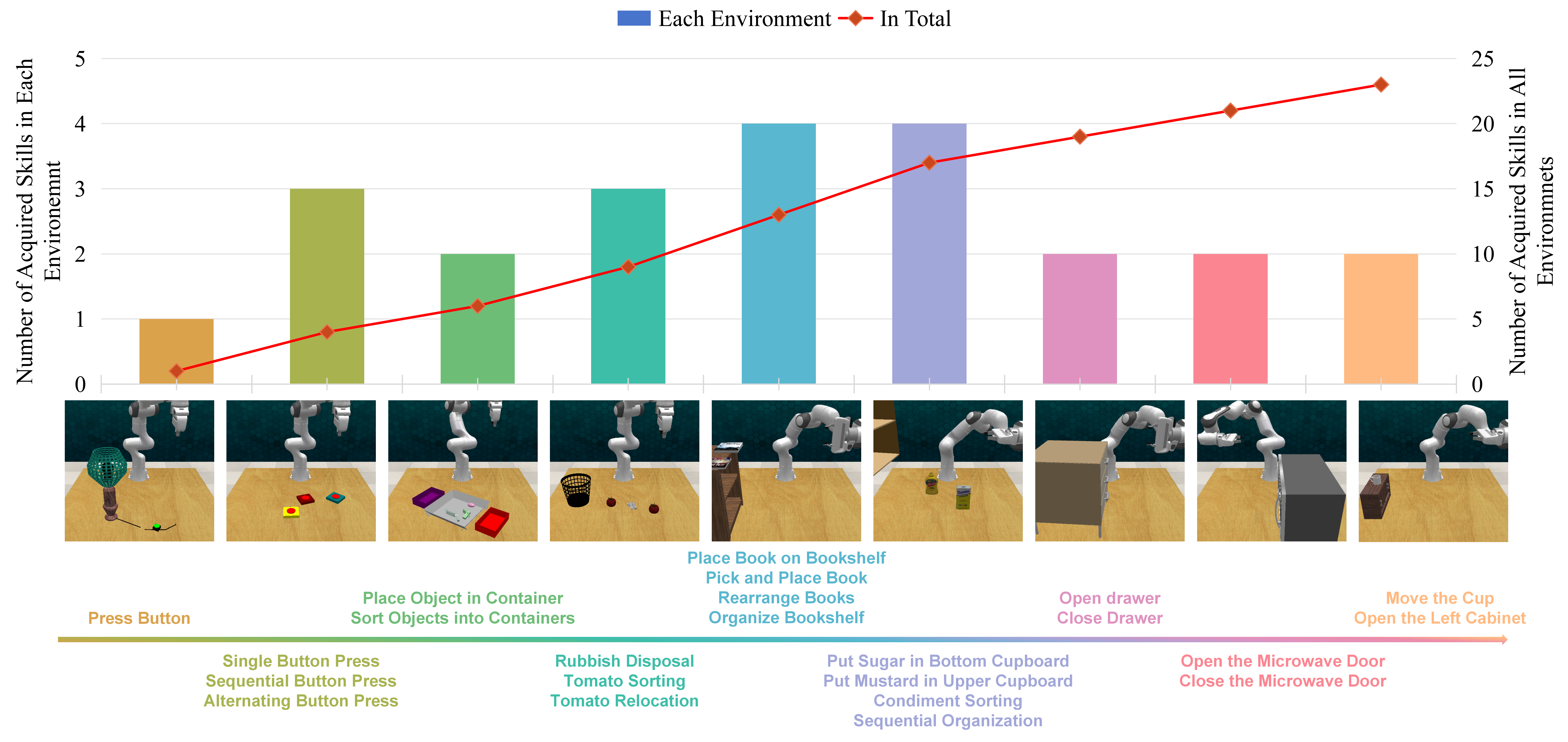}
\caption{\textbf{Skills acquirement process in RLBench.} The number of acquired skills increases as the robot explores different environments.}
\label{figure exploration in rlbench}
\end{figure*}

\subsection{Everyday Robotics Tasks} \label{section rlbench experiment}

To answer the second question, we conduct a series of experiments in RLBench \cite{james2020rlbench} to evaluate the generalization ability of GExp. As shown in Figure \ref{figure scenes in rlbench}, nine different scenes are chosen, including Turning on a Lamp, Pressing Buttons, Moving Object to Containers, Clearing Rubbish, Arranging Bookshelf, Arranging Cupboard, Manipulating Drawer, Manipulating Microwave, and Placing Cup to Cabinet. The detailed information of different scenes is listed in Appendix \ref{appendix scenes rlbench}.

We first let robots interact with the environments and acquire skills through self-exploration. We arrange 9 environments above according to manipulation complexity and record the number of acquired skills in each environment and the total number of acquired skills across all environments. As is demonstrated in Figure \ref{figure exploration in rlbench}, after the exploration phase, the robot manages to acquire 1 to 4 skills in each environment, with a total of 23 skills across all environments. This result shows that GExp can drive robots to acquire skills efficiently by exploration. Furthermore, we find that there is a relation between the number of acquired skills and the number of objects in the environment, partially because foundation models tend to generate sequential tasks that involve multiple objects.

To evaluate the quality of collected skills in GExp, we set up 2 complex tasks: Desktop Organization and Cup Acquisition. Desktop Organization involves many objects and needs relatively long-horizon planning. Cup Acquisition only involves two objects but the manipulation steps are highly correlated, which means that the success of subsequent actions depends on the success of the current action. More details about the environments can be found in Appendix \ref{appendix scenes rlbench}. We compare GExp with a baseline that is the same as GExp but without backtracking control to show the importance of closed-loop control. We employ the acquired skills to complete these 2 tasks and evaluate the success rate. Moreover, the two tasks can be divided into several sub-tasks, and we also report the success rate of the sub-tasks. 

As is shown in Table \ref{tab rlbench complex}, GExp outperforms baseline in 2 tasks and almost all sub-tasks, especially in difficult sub-tasks such as Clear Rubbish and Close Drawer. During the experiment, we found that backtracking control could detect failure and retry the task, which significantly improves the successful rate of completing the whole task. This is particularly beneficial to Cup Acquisition since this task is highly correlated. We show some backtracking control cases in Appendix \ref{appendix results rlbench}. There are also some failures. For example, in Cup Acquisition, the drawer is regarded as a table sometimes, causing failures in self-verification. It explains why the successful rate of Pick\&Place Cup is only 0.5 when using backtracking control. With the development of foundation models, we expect this problem to be mitigated. We include more details in Appendix \ref{appendix results rlbench}.

\begin{table}[ht]
\centering
\begin{tabular}{lcc}
\toprule
\textbf{Task}                       & \textbf{Open-Loop} & \textbf{Backtracking} \\
\midrule
\textbf{Desktop Organization}      & 0.55              & \textbf{0.85}                \\
Clear Rubbish                & 0.60               & \underline{0.90}               \\
Place Tomato1                & 0.85              & \underline{0.95}                 \\
Place Tomato2                & 0.95              & \underline{1.00}                 \\
Place Chocolate Jello        & 0.90               & 0.90                 \\
Place Strawberry Jello       & 1.00                 & 1.00                 \\
\hline
\textbf{Cup Acquisition}           & 0.20               & \textbf{0.45}                \\
Open Drawer                  & 0.65              & \underline{0.90}               \\
Pick\&Place Cup              & 0.25              & \underline{0.50}                 \\
Close Drawer                 & 0.60               & \underline{0.85}                 \\

\bottomrule
\end{tabular}
\caption{\textbf{Open-loop vs. backtracking.} We evaluate the performance of robot real-solution on two specially designed tasks. Results show that the proposed backtracking method can significantly improve the successful rate of execution.}
\label{tab rlbench complex}
\end{table}

\subsection{Real-world Deployment}

\begin{figure*}[h]
\centering
\includegraphics[width=1\textwidth]{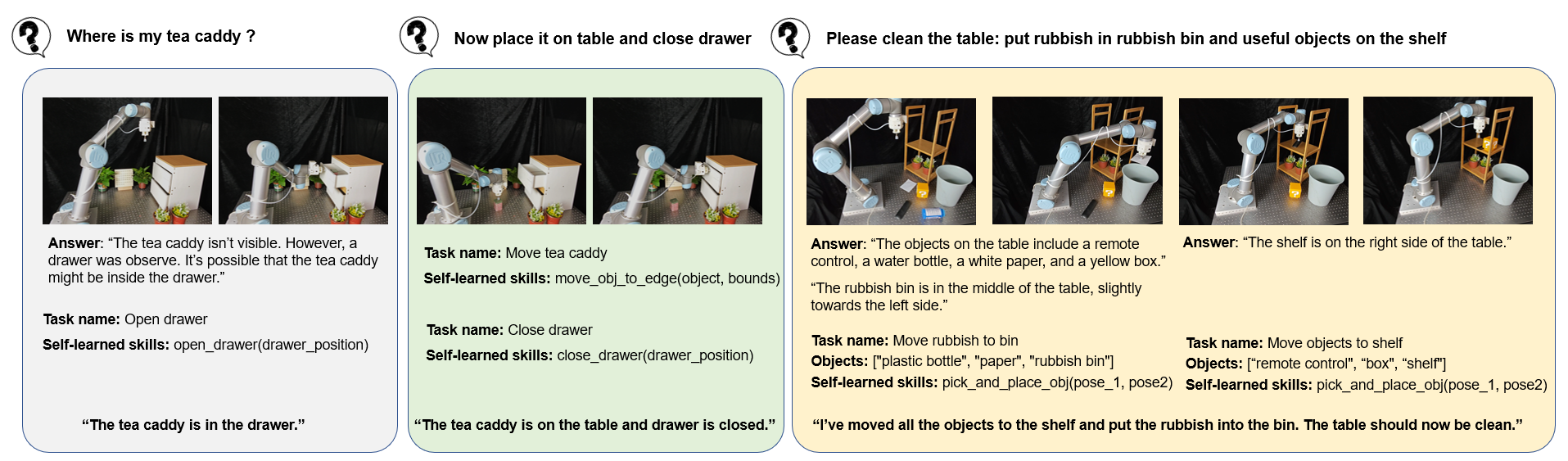}
\caption{\textbf{Skills acquirement process in RLBench.} The number of acquired skills increases as the robot explores different environments.}
\label{figure exploration in rlbench}
\end{figure*}

For the last question, we set up a real-world tabletop environment letting robot letting robots use a self-learned skills library to accomplish user instruction. As shown in Figure \ref{Figure real world experiment}, a UR5 robot arm with 6-DoF equipped with a suction gripper is placed on the table, which can be controlled by using skills learned in Section \ref{section rlbench experiment}. We use a RealSence D415 for perception. The captured RGB image can be used for scene understanding (realized by GPT4V) and object position acquisition (realized by VLMs including ViLD \cite{gu2021open} and Segment Anything \cite{kirillov2023segment}).

We sim-to-real two evaluation tasks introduced in Section \ref{section rlbench experiment}: Desktop Organization and Cup Acquisition. Different from directly providing a specific task in a simulation environment, a free-form language instruction is given which needs to be transferred into tasks using the workflow introduced in Section \ref{section deployment}. To further evaluate the generalization ability of GExp, we use objects that are different from those learned in the exploration. For instance, the cup is replaced by a tea caddy, and the objects in the clean table task are a remote controller, plastic bottle, paper, and toy box. Based on the results, we find the proposed framework can successfully transfer user instruction to feasible tasks. The skills can be chosen correctly from the self-learned library and realized the task successfully. The commonsense knowledge of GPT4V also empowers the self-learned skills generalization ability, where they can be used to manipulate objects not seen before. The experiment log can be found in Appendix \ref{appendix real-world find cup} and \ref{appendix real-world clean table}. The videos of real-world experiments can be found on our project website: \href{https://marcyu0303.github.io/GExp.github.io/}{\textcolor{blue}{GExp.github.io}}.

\section{Conclusion, Limitations and Future Work}
In this paper, we propose GExp, a novel framework for robot self-exploration utilizing the power of foundation models, which enables robots to explore and learn autonomously without human intervention. To achieve this goal, we devise modules including self-exploration, knowledge-base-building, and close-loop feedback based on foundation models. To validate the effectiveness of the framework, we designed a series of simulation experiments and deployed the framework on a real robot platform. Our results show that GExp allows robots to explore autonomously and learn from exploration by generating useful skills. Based on this framework robots can learn autonomously without relying on human guidance, which is a milestone in the development of robotic intelligence

The main limitation of our method is reliance on the power of foundation models. We find that there still exists a gap between the current ability of foundation models and the requirement of designing a robust and stable system based on them for robotics tasks. GExp is one attempt to implement foundation models enhancing autonomous robots. In the future, we will try to combine learning-based methods into this framework for low-level robot control. We believe the further development of foundation models could further improve the performance of GExp, allowing it to be applied to a wider range of robotics tasks.

\newpage
\newpage
\bibliographystyle{named}
\bibliography{references}
\newpage
\appendix

\section{Modules in GExp}
\subsection{How GExp understand the environment?} \label{appendix section 1}

\begin{tcolorbox}
[breakable, colback=Emerald!10,colframe=cyan!40!black,title=\textbf{Prompt of scene understanding}]

You will be given a picture of table-top scene observation of robot system. You need to give a description about the picture. The description should contain following information:

* What objects are on the table. Please describe those objects. \newline
* You do not need to describe the robot arm, please focus on other objects you observed.

You should output your answer in a json format. The output format should be:
\begin{lstlisting}[language=json]
```json
{
"Description": "Describe what you have observed on the table.",
"Objects on table":[
    {
        "Name": "The name of object",
        "Color": "object's color",
    },
    ]
}
```
\end{lstlisting}
\end{tcolorbox}

\begin{figure}[htbp]
\centering
\includegraphics[width=0.4\textwidth]{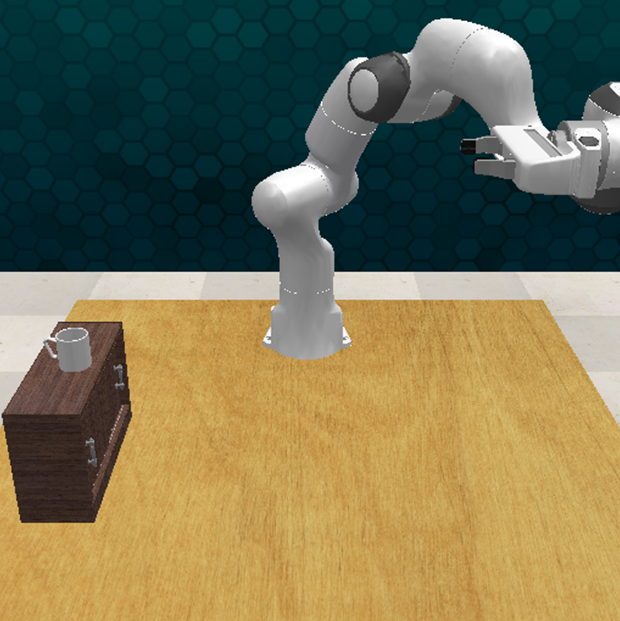}
\caption{Initial observation and current observation.}
\end{figure}

\begin{tcolorbox}
[breakable, colback=Emerald!10,colframe=cyan!40!black,title=\textbf{Output of VLM}]
\begin{lstlisting}[language=json]
```json
{
  "Description": "On the table, there is a wooden brown drawer cabinet situated on the left side. Visual center of the table is almost empty except the robot system which is not the point of interest. Directly on the top side of the drawer, there is a white cup with a handle facing towards the drawer.",
  "Objects on table": [
      {
          "Name": "Drawer cabinet",
          "Color": "Brown"
      },
      {
          "Name": "Cup",
          "Color": "White"
      }
  ]
}
```
\end{lstlisting}
\end{tcolorbox}
\subsection{How GExp generate tasks based on observation?} \label{appendix prompt generate tasks}
\begin{tcolorbox}
[breakable, colback=Emerald!10,colframe=cyan!40!black,title=\textbf{Prompt of tasks generation module}]
You are a helpful AI assistant. You are helping to create ten different tabletop manipulation tasks according to the provided information.The generated tasks should in the order from easy to hard. The first task should be the easiest one which requires a small number of steps to realize it. The design of the latter task needs to be based on the former. The design purpose of tasks gradually changed from testing basic manipulation ability to long-range planning ability. \newline
The output format should be in the form of json: \newline
\begin{lstlisting}[language=json]
```json
{
    "Task Name": "The name of generated task",
    "Objects": ["A list of involved objects"],
    "Task Description": "Necessary description of the task"
}
```
\end{lstlisting}

Rules: \newline
* You must write the full name of the object, not just its color. \newline
* Don't use 'all objects', you should write them one by one. \newline
* The task should involved robot manipulation not just observation. \newline
* Do not create task with any assume value of distance or position, you can use relative value to the boundary. \newline
* You should output ```json and ``` showing the output is json block.
\end{tcolorbox}

\subsection{How GExp realize planning and controlling the robot?} \label{appendix prompt planning}

\begin{tcolorbox}
[breakable, colback=Emerald!10,colframe=cyan!40!black,title=\textbf{Prompt of planning module}]
You are a sophisticated AI assistant tasked with creating a multi-step plan for a robot manipulation task. Your role is to design this plan methodically, ensuring accuracy and precision. To successfully execute this task, please adhere to the following guidelines and rules: \newline \newline
* Library Restrictions: The only permitted library is `numpy`. Please import it as `np` if necessary. Avoid using any other libraries. \newline
* Use Existing Functions: Do not create new functions. Only utilize the functions that are provided. \newline
* Variable Constraints: Avoid using variables that require user specification or complex calculations. Instead, derive exact values based on the given variables. \newline
* Position and Size Considerations: Do not make assumptions about positions or sizes. \newline
* Adherence to BOUNDS: Ensure that all positions and movements fall within the robot arm's working BOUNDS. When establishing new positions, reference the BOUNDS information for accuracy. \newline
* Completeness of Plan: You should output the whole plan without using the expression like "repeat steps 2-5 to ...". \newline
* Efficient Function Usage: Some high-level functions are provided to help you finish the task. You should try to use them to optimize the plan. \newline \newline
Format for the Plan: \newline
Your plan should be structured in a series of JSON blocks, each representing a step in the process. The format for each block is as follows: \newline
\begin{lstlisting}[language=json]
```json
{
    "Name": "Provide a name for this step.",
    "Explanation": "Include a necessary explanation about this step.",
    "Code": "Write the Python code to execute this step. You should write the code in a single line using '\\n' for changing line." 
}
```
\end{lstlisting}
*Note:
The number of JSON blocks should correspond to the number of steps in your plan.
You should output ```json and ``` showing the output is json block.

Please remember that you can only use following defined functions:
\end{tcolorbox}

\subsection{How GExp verify the precondition or task success or not with VLM?} \label{appendix prompt verify}
\begin{tcolorbox}
[breakable, colback=Emerald!10,colframe=cyan!40!black,title=\textbf{Prompt of verification module}]
You are a helpful AI assistant. Your job is to determine if a condition is satisfied or not. You will be given a description of the condition and an RGB image capturing the current situation. Please determine it in a step by step way to be sure that we have the right answer. 

To help you understand the scene, I will give you a description of the initial scene setup and a corresponding image. So you will get two images, the first is the initial scene image, and the second is the current scene image.
\begin{lstlisting}[language=json]
```json
{
    "condition": "The box is grasped by the robot arm",
    "initial description": f"{VLM_initial_observation}",
}
```
\end{lstlisting}
\end{tcolorbox}

\begin{figure}[htbp]
\centering
\includegraphics[width=0.48\textwidth]{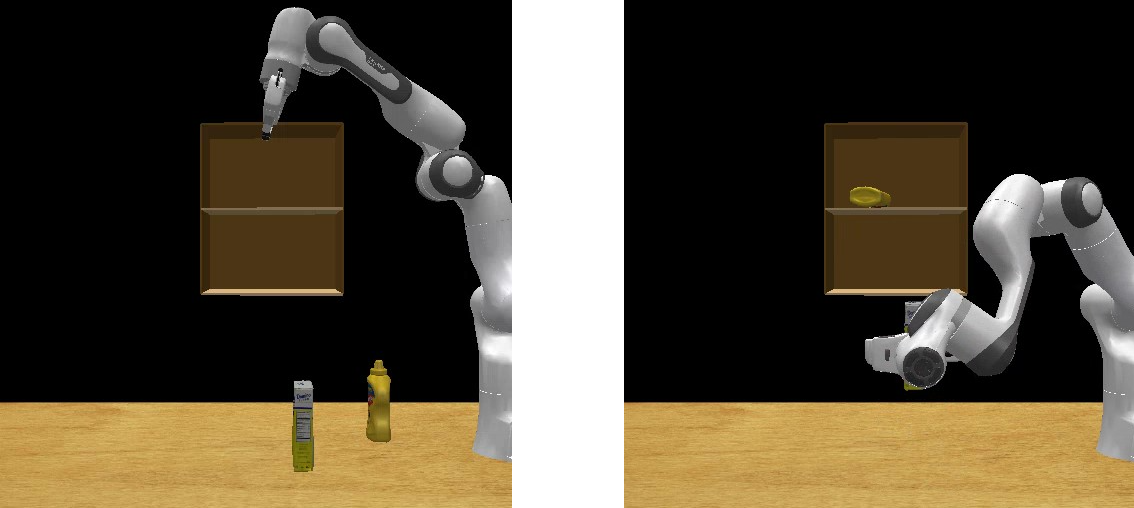}
\caption{Initial observation and current observation of cabinet environment. The verify condition is whether the robot arm has grasped the milk box.}
\end{figure}

\begin{tcolorbox}
[breakable, colback=Emerald!10,colframe=cyan!40!black,title=\textbf{Output of VLM}]
To determine if the condition "The box is grasped by the robot arm" is satisfied, let's consider the two images provided:

1. Initial scene description image: This image displays the initial setup with the cupboard, the robot arm, the table, the box, and the yellow bottle. The box is on the table, to the left of the yellow bottle, and not being grasped by the robot arm, which is positioned to the right side of the box, with the gripper open and not in contact with the box.

2. Current scene description image: This second image shows a different state where the robot arm has moved from its initial position, and the end-effector (gripper) of the robot arm appears to be surrounding the box, indicating that the arm has attempted to grab it. The yellow bottle is within the box, suggesting some movement of objects has occurred. It is not possible to confirm a firm grasp visually without additional cues like the arm's grip pressure or the box's weight distribution. However, the change in position coupled with the gripper's placement around the box strongly suggests that the robot arm is now holding the box.

Given this analysis, I can conclude that the condition "The box is grasped by the robot arm" is met.
\end{tcolorbox}

\subsection{How GExp generate skill from a successful experience?} \label{appendix prompt learn skill}
\begin{tcolorbox}
[breakable, colback=Emerald!10,colframe=cyan!40!black,title=\textbf{Prompt of skill generation}]
You are a helpful AI assistant expertise in programming robot controlling code. You will be given with a Python script which is used to solve a robot manipulation problem. Your task is creating one Python function that helps people in future solve similar task in an easier way. Your function will be send into function library for further use. \newline \newline
Following rules you should follow when creating new function: \newline
* Library Restrictions: The only permitted library is `numpy`. Please import it as `np` if necessary. Avoid using any other libraries. \newline
* Use Existing Functions: When creating new function, you should use existing functions in library. \newline
* Function should be general: The generated new function should be general. Do not generate a function to realize the given specific task. \newline \newline
Format of your response: \newline
You should output a json block as your output, which follows the following format: \newline
\begin{lstlisting}[language=json]
```json
{
    "Function name": "Name_of_your_function(), for example lift_object().",
    "Description": "Describe the function your created.",
    "Input": "What is the input of your function.",
    "Output": "What is the output of your function.",
    "Code": "The code of generated function.",
    "Example": "You should rewrite the code using new function.",
    "Related functions": "what functions are used in this function."
}
```
\end{lstlisting}
Existing functions in function library: \newline f"\{existing\_skills\_library\}"
\end{tcolorbox}

\section{"BLOCKS WORLD" Experiment}
\subsection{Tasks generated in "BLOCKS WORLD"} \label{appendix tasks in blocks world}
\begin{tcolorbox}[breakable, colback=Emerald!10,colframe=cyan!40!black,title=\textbf{Output of tasks generate modul}]
\begin{lstlisting}[language=json]
{
    {
        "Task Name": "Pick and Place the Purple Block",
        "Objects": [
          "purple block"
        ],
        "Task Description": "The robot arm should identify the purple block, pick it up, and place it at the right boundary of the table."
    },
    {
        "Task Name": "Create a Two-Block Stack",
        "Objects": [
          "purple block",
          "blue block"
        ],
        "Task Description": "The robot needs to stack the purple block on top of the blue block."
    },
    {
        "Task Name": "Create a Three-Block Stack",
        "Objects": [
          "purple block",
          "blue block",
          "green block"
        ],
        "Task Description": "The robot arm should pick up the green block and place it on top of the stack of the purple and blue blocks."
    },
    {
        "Task Name": "Color Match and Stack",
        "Objects": [
          "purple block",
          "blue block",
          "green block",
          "yellow block",
          "orange block",
          "red block"
        ],
        "Task Description": "The task is to create three stacks of two blocks each, where each stack contains blocks of complementary colors. For example, one stack could have the purple and yellow blocks, the second could have the blue and orange blocks, and the third could have the green and red blocks."
    },
    {
        "Task Name": "Block Pyramid Stacking",
        "Objects": [
          "yellow block",
          "orange block",
          "red block"
        ],
        "Task Description": "The robot needs to stack the blocks into a pyramid shape in the center of table. The bottom layer would consist of the yellow and orange blocks side by side, and the red block should be placed on top."
    },
    {
        "Task Name": "Complex Pyramid Construction",
        "Objects": [
          "purple block",
          "blue block",
          "green block",
          "yellow block",
          "orange block"
        ],
        "Task Description": "The robot needs to stack all the blocks into a pyramid with three levels in the center of table. The bottom level should consist of the purple, blue, and green blocks. The middle level should have the yellow and orange blocks. The red block should be on top. The robot needs to plan the sequence and positions of its movements carefully to build the pyramid without knocking any blocks over."
    }
}
\end{lstlisting}
\end{tcolorbox}
\subsection{Skills generated in "BLOCKS WORLD"} \label{appendix skills in blocks world}

\begin{tcolorbox}[breakable, colback=Emerald!10,colframe=cyan!40!black,title=\textbf{Conversation log of finding cup task}]
\begin{lstlisting}[language=json]
{
  "pick_and_place_object(object_name, destination)": {
    "Type": "function",
    "Description": "This function is used to pick and place a specific object from its current location to a specified destination. The robot identifies the object, picks it up, and places it at the destination.",
    "Input": "('object_name': str, 'destination': list). 'object_name' is the name of the object to be moved, 'destination' is a list of 3 elements representing the coordinates where the object will be placed.",
    "Output": "None. The function only performs actions, it does not return any value.",
    "Related functions": "get_obj_position(), get_obj_dimensions(), movep(), close_gripper(), open_gripper(), go_home()",
    "Code": "def pick_and_place_object(object_name, destination): \\n  object_position = get_obj_position(object_name) \\n  object_dimensions = get_obj_dimensions(object_name) \\n  BOUNDS = [[0.25, 0.75],[-0.5, 0.5],[0.0, 0.28]] \\n  movep((object_position[0], object_position[1], object_position[2] + object_dimensions[2])) \\n  movep((object_position[0], object_position[1], object_position[2])) \\n  close_gripper() \\n  movep((object_position[0], object_position[1], BOUNDS[2][1])) \\n  movep((destination[0], destination[1], BOUNDS[2][1])) \\n  movep((destination[0], destination[1], BOUNDS[2][0] + destination[2])) \\n  open_gripper() \\n  go_home() \\n "
  },
  "stack_blocks(block1, block2)": {
    "Type": "function",
    "Description": "This function is designed to stack a specific block on top of another block. It uses robot functions such as 'go_home()', 'pick_and_place_object()', 'get_obj_position()' and 'get_obj_dimensions()' to achieve this task. This function will help automate the process of block stacking in future robot manipulation tasks.",
    "Input": "Two strings, block1 and block2, which represent the names of the blocks to be stacked.",
    "Output": "None. The function performs the action of moving the robot to stack the blocks but doesn't return anything.",
    "Related functions": "go_home(), pick_and_place_object(), get_obj_position(), get_obj_dimensions()",
    "Example": "stack_blocks('purple block', 'blue block')",
    "Code": "def stack_blocks(block1, block2):\\n    go_home()\\n    pick_and_place_object(block1, list(np.array(get_obj_position(block2)) + np.array([0, 0, get_obj_dimensions(block2)[2]])))\\n    go_home()"
  },
  "create_three_block_stack()": {
    "Type": "function",
    "Description": "This function is used to create a three-block stack. It first stacks the first two blocks, then calculates the position to place the third block, moves the third block to the stack, and finally, the robot returns to the home position.",
    "Input": "Three parameters: first two blocks for initial stacking and the third block to be placed on top of the stack.",
    "Output": "The function does not return any output. Its purpose is to perform the action of creating a three-block stack.",
    "Related functions": "stack_blocks(), get_obj_position(), get_obj_dimensions(), pick_and_place_object(), go_home()",
    "Example": "Example usage: create_three_block_stack('purple block', 'blue block', 'green block')",
    "Code": "def create_three_block_stack(first_block, second_block, third_block): \\n    # Step 1: Stack first and second Blocks \\n    stack_blocks(first_block, second_block) \\n    # Step 2: Get Position and Dimensions of the Stack \\n    stack_position = get_obj_position(second_block) \\n    stack_dimensions = get_obj_dimensions(second_block) \\n    # Step 3: Calculate Position to Place third Block \\n    third_block_position = (stack_position[0], stack_position[1], stack_position[2] + stack_dimensions[2]/2 + get_obj_dimensions(third_block)[2]/2) \\n    # Step 4: Move third Block to the Stack \\n    pick_and_place_object(third_block, third_block_position) \\n    # Step 5: Return to Home Position \\n    go_home()"
  },
  "build_pyramid(blocks_order)": {
    "Type": "function",
    "Description": "This function is designed to build a pyramid shape with the blocks. It takes an array of block names in the order from bottom to top. The first two blocks will be placed side by side at the bottom. The next block will be placed on top of the first two blocks.",
    "Input": "An array of block names, from bottom to top. For example, ['yellow block', 'orange block', 'red block'] means yellow and orange blocks will be placed at the bottom and red block on top.",
    "Output": "No output. The function commands the robot to build the pyramid according to the block order.",
    "Related functions": "Functions used in this function are: pick_and_place_object(), get_obj_position(), get_obj_dimensions(), go_home().",
    "Example": "Example of usage: build_pyramid(['yellow block', 'orange block', 'red block'])",
    "Code": "def build_pyramid(blocks_order): \\n  # Step 1: Move first block to the right boundary \\n  pick_and_place_object(blocks_order[0], [BOUNDS[0][1], BOUNDS[1][0], BOUNDS[2][0]]) \\n  # Step 2: Move second block next to the first block \\n  first_block_position = get_obj_position(blocks_order[0]) \\n  first_block_dimensions = get_obj_dimensions(blocks_order[0]) \\n  second_destination = [first_block_position[0] + first_block_dimensions[0], first_block_position[1], first_block_position[2]] \\n  pick_and_place_object(blocks_order[1], second_destination) \\n  # Step 3: Stack third block on top of the first and second blocks \\n  second_block_position = get_obj_position(blocks_order[1]) \\n  third_destination = [(first_block_position[0] + second_block_position[0]) / 2, first_block_position[1], first_block_position[2] + first_block_dimensions[2]] \\n  pick_and_place_object(blocks_order[2], third_destination) \\n  # Step 4: Return to home position \\n  go_home()"
  }
}
\end{lstlisting}
\end{tcolorbox}

\newpage
\section{RLBench Experiment}

\subsection{RLBench-based environments} \label{appendix scenes rlbench}





\begin{figure}[htbp]
\centering
\includegraphics[width=0.45\textwidth]{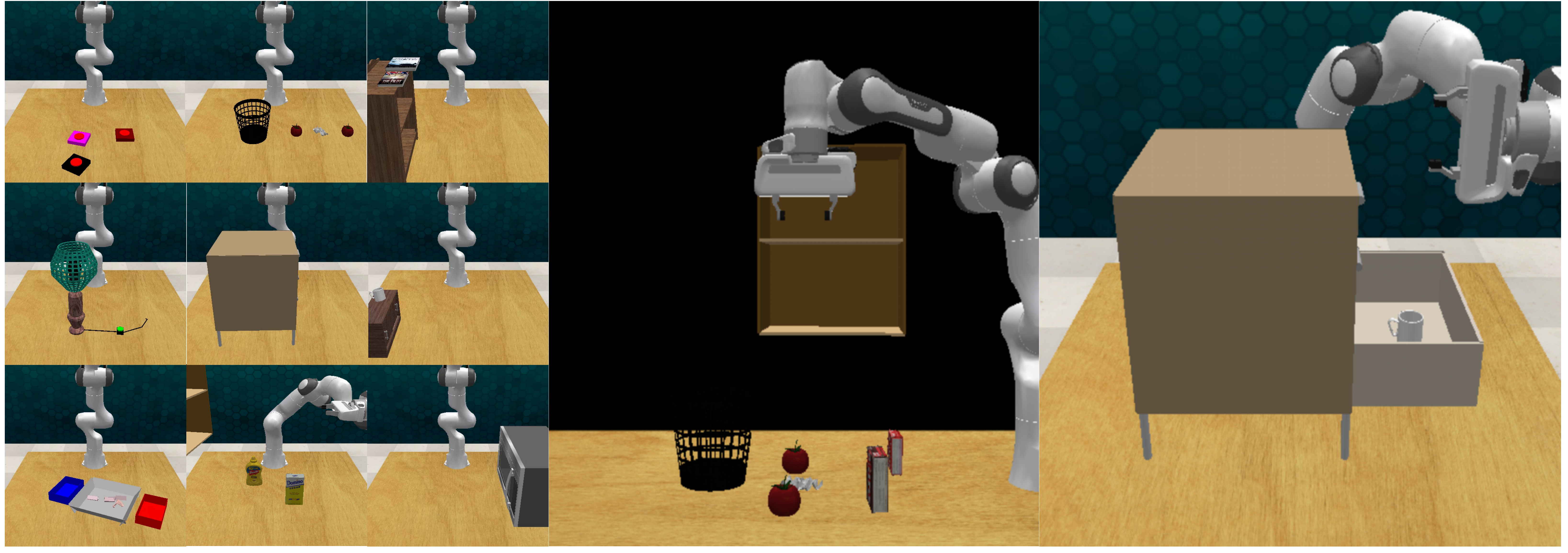}
\caption{\textbf{Environments in RLBench.} Left are 9 environments for exploration and skill acquisition. Middle and right are environments of Desktop Organization and Cup Acquisition respectively.}
\label{figure scenes in rlbench}
\end{figure}

To test the capability of GExp in daily tasks, we first utilize nine environments in RLBench, which are shown in Figure \ref{figure scenes in rlbench}.

     \textit{Turning on a Lamp}: There are a lamp and a button on the table. Robot can turn the lamp on by pressing the button.    
     
     \textit{Pressing Buttons}: There are 3 buttons on the table. Robot can press them sequentially. This process can be seen in daily tasks, such as microwaving food. This environment is more complex than Turning on a Lamp. 
     
     \textit{Moving Object to Containers}: One large container and two small containers are on the table. Several objects are in the larger container. Robot can explore and acquire some basic picking or placing skills in this environment.   
     
     \textit{Clearing Rubbish}: One rubbish bin, two tomatoes and one paper rubbish are on the table. Robot is hoped to clear the rubbish by picking and placing it into the bin.

     \textit{Arranging Bookshelf}: A bookshelf is placed on the table. Two books are on the top of the bookshelf. Robot can try to place them in the bookshelf.

     \textit{Arranging Cupboard}: A sugar box and a milk bottle are on the table. A cupboard is near the table. The sugar box and the milk bottle can be transferred into the cupboard by robot.

     \textit{Manipulating Drawer}: A three-layer drawer is put on the table. Different from above environments, this environment involves articulated object.

     \textit{Manipulating Microwave}: A microwave is placed on the table. Robot can try to open and close it.

     \textit{Placing Cup to Cabinet}: A cabinet and a cup are on the table. The cabinet has two doors, which can be opened by sliding in left or right direction. This task combines articulated objects with non-articulated objects.

To further prove the importance of acquired skills and backtracking control, we also build two complex environments and tasks. Environments are shown in Figure \ref{figure scenes in rlbench}.

    \textit{Desktop Organization}: We combine \textit{Clear Rubbish} with \textit{Arranging Cupboard}. A paper rubbish, a rubbish bin, two tomatoes, a chocolate jello box and a strawberry jello box are placed on the table. A cupboard is positioned nearby. In this task, we require robot to clean rubbish and organize other objects to proper positions. This task mainly tests the long-horizon planning and skill usage of GExp. 

    \textit{Cup Acquisition}: There are a cup and drawer in this scene. The cup is placed in the bottom drawer and is not observable. We require robot to find the cup by opening drawer and place the cup on the table. In the end, robot should close the drawer. This task is especially difficult since the steps are highly correlated, as mentioned in \ref{section rlbench experiment}.

\subsection{Specific cases in RLBench-based environments} \label{appendix results rlbench}
In this section, we provide some cases on backtracking-based close-loop control and failures.

\begin{figure}[htbp]
\centering
\includegraphics[width=0.5\textwidth]{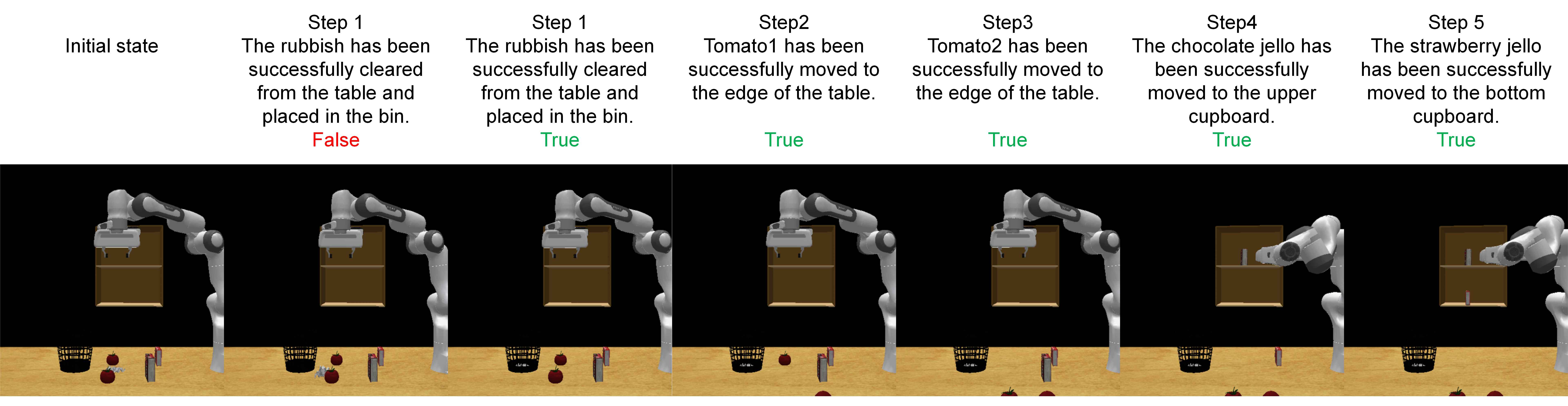}
\caption{\textbf{Close-loop case in Desktop Organization task.} Robot manages to detect failure and retry clearing rubbish.}
\label{figure desktop organization}
\end{figure}

\begin{figure}[htbp]
\centering
\includegraphics[width=0.5\textwidth]{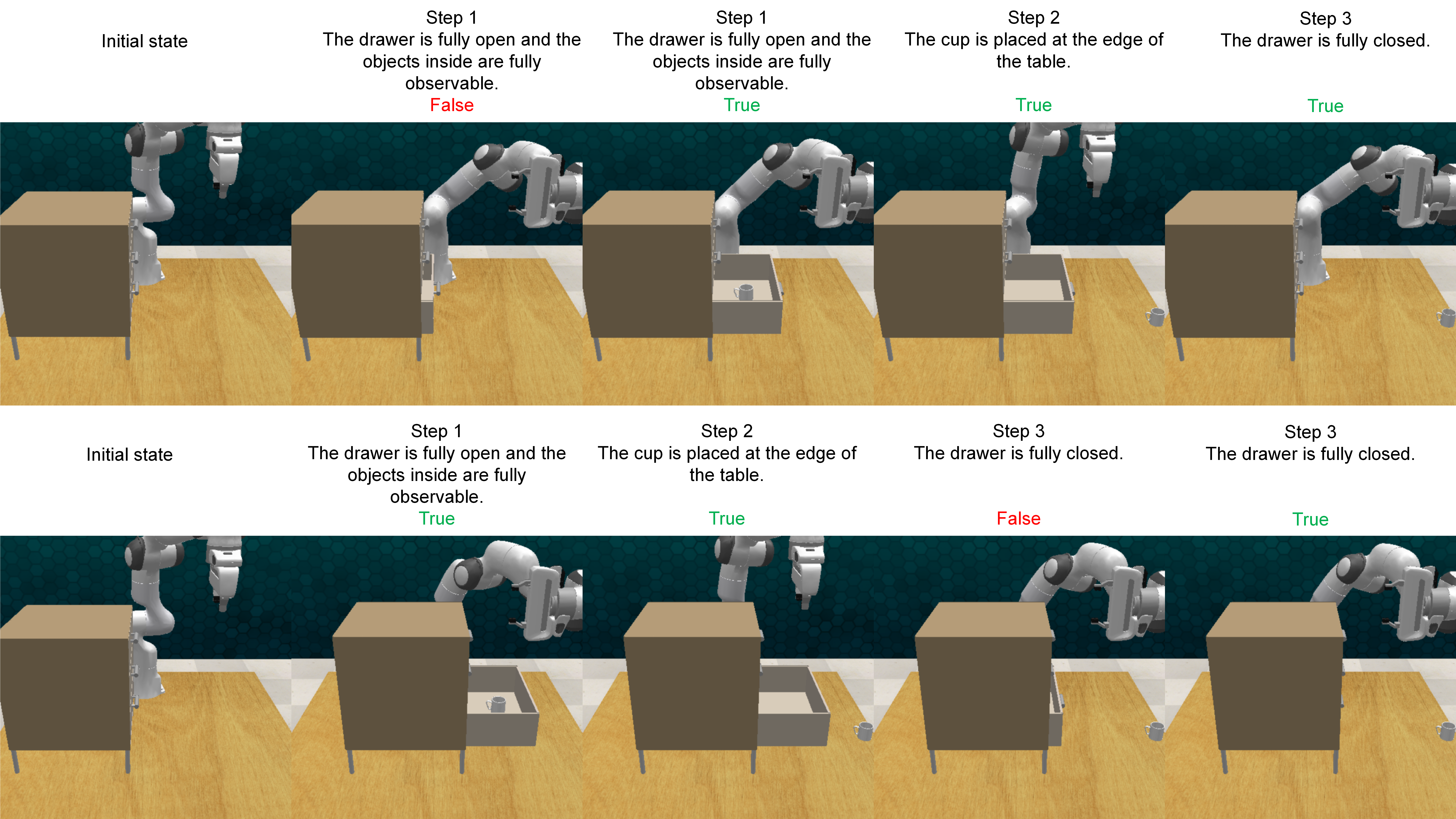}
\caption{\textbf{Close-loop cases in Cup Acquisition task.} Robot manages to reopen the drawer and re-close the drawer.}
\label{figure cup acquisition}
\end{figure}

As seen in Figure \ref{figure desktop organization}, robot fails to place the rubbish in the rubbish bin due to unstable grasping. However, with VLM as a verifier and the precondition generated by LLM, robot can detect the failure and retry to clear the rubbish. Some similar cases can be also observed in Cup Acquisition task. For example, as is shown in Figure \ref{figure cup acquisition}, robot does not open the drawer fully, close the drawer, or loosens the cup occasionally, causing the rest of task unfinished. Backtracking-based close-loop control endows robot to reflect and adjust according to current state. 

\begin{figure}[htbp]
\centering
\includegraphics[width=0.5\textwidth]{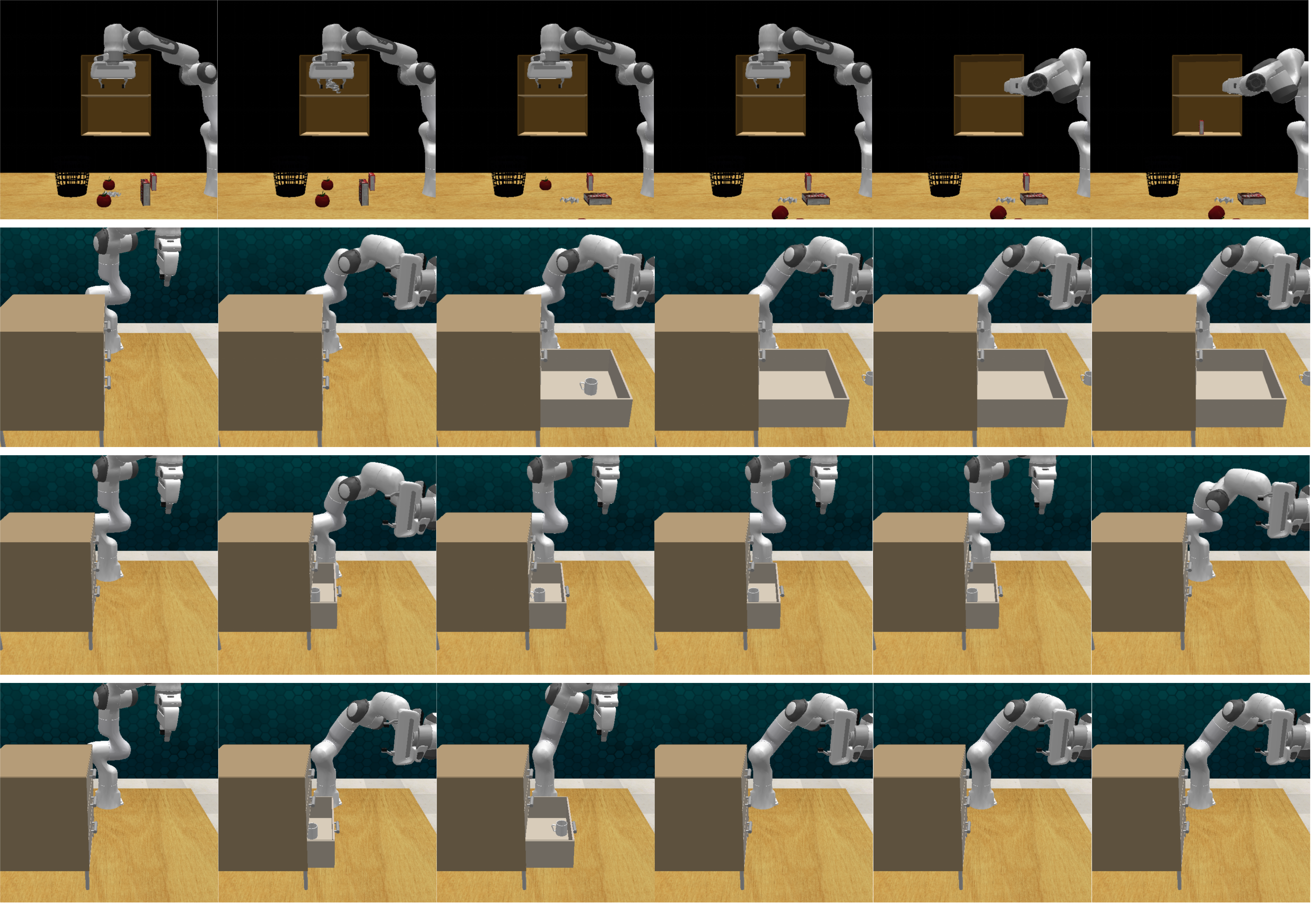}
\caption{\textbf{Failure cases.} \textbf{First row}: the falling rubbish knocks down a box, resulting ungraspable pose for robot. \textbf{Second row}: the drawer is opened so large that robot can not fetch the handle(handle is not in the reachable space of robot arm). \textbf{Third row}: robot thinks the drawer is open but it is not opened fully so robot can not pick the cup up. \textbf{Fourth row}: robot regards the drawer as a table, not realizing the failure of placing cup.}
\label{figure failure cases}
\end{figure}

Nevertheless, GExp also fails in some cases, which are illustrated in Figure \ref{figure failure cases}. In Desktop Organization,  one of boxes is knocked over due to unstable grasping. This results ungraspable poses. Therefore, in the subsequent steps, robot fails to pick up the box. In Cup Acquisition Organization, failures occurs more frequently, particularly in opening drawer and picking up cup. When robot executes these two steps, there is dilemma that the drawer is partially open but robot can not pick it, as is shown in the third row of Figure
\ref{figure failure cases}. Moreover, VLM malfunctions sporadically, regarding the drawer as a table. And randomness in manipulation will also cause unrecoverable states.

\newpage
\section{Real-world Experiments}
\subsection{Finding the cup} \label{appendix real-world find cup}
\begin{tcolorbox}[breakable, colback=Emerald!10,colframe=cyan!40!black,title=\textbf{Conversation log of finding cup task}]
\begin{lstlisting}[language=json]
User:
I can't find my tea caddy.
\end{lstlisting}

\begin{lstlisting}[language=json]
Controller:
{
    "Thought": "The user can't find their tea caddy. The first step is to observe the environment to locate the missing item.",
    "Action": "observe()",
    "Action input": {"query": "Where is the tea caddy?"}
}
\end{lstlisting}

\begin{lstlisting}[language=json]
Observer:
{
    "answer": "The tea caddy is not visible in the picture.",
    "observed_objs": ["drawer"]
}
\end{lstlisting}

\begin{lstlisting}[language=json]
Controller:
{
    "Thought": "The tea caddy isn't currently visible. However, a drawer was observed in the environment. It's possible that the tea caddy might be inside the drawer.",
    "Action": "observe()",
    "Action input": {"query": "What is inside the drawer?"}
}
\end{lstlisting}

\begin{lstlisting}[language=json]
Observer:
{
    "answer": "The image does not show the inside of the drawer, so it is impossible to determine what is inside it.",
    "observed_objs": ["drawer"]
}
\end{lstlisting}

\begin{lstlisting}[language=json]
Controller:
{
    "Thought": "Since I can't see inside the drawer, the next logical step would be to open it and check if the tea caddy is inside.",
    "Action": "execute_task()",
    "Action input": {"Task name": "Open the drawer", "Objects": ["drawer"], "Task description": "The robot arm needs to open the drawer to check if the tea caddy is inside."}
}
\end{lstlisting}

\begin{lstlisting}[language=json]
Executor:
    "open_drawer('drawer')"
\end{lstlisting}

\begin{lstlisting}[language=json]
Observer:
{
    "Thought": "The drawer has been opened successfully. Now I need to observe the contents of the drawer to see if the tea caddy is inside.",
    "Action": "observe()",
    "Action input": {"query": "What is inside the drawer now?"}
}
\end{lstlisting}

\begin{lstlisting}[language=json]
Observer:
{
    "answer": "The tea caddy is in the drawer.",
    "observed_objs": ["drawer", "tea caddy"]
}
\end{lstlisting}

\begin{lstlisting}[language=json]
Controller:
{
    "Thought": "The tea caddy has been observed inside the drawer. Now I should inform the user about the location of the tea caddy.",
    "Action": "finish()",
    "Action input": {"message": "I found your tea caddy. It is in the drawer."}
}
\end{lstlisting}

\begin{lstlisting}[language=json]
user:
Now the drawer is open, you don't need to open again. Move tea caddy to edge of table and close drawer.
\end{lstlisting}

\begin{lstlisting}[language=json]
Controller:
{
    "Thought": "The user wants me to move the tea caddy to the edge of the table and then close the drawer. I will first move the tea caddy to the specified location.",
    "Action": "execute_task()",
    "Action input": {"Task name": "Move the tea caddy", "Objects": ["tea caddy"], "Task description": "The robot arm should move the tea caddy to the edge of the table."}
}
\end{lstlisting}

\begin{lstlisting}[language=json]
Executor:
    "move_object_to_edge('tea caddy', BOUNDS)"
    "close_drawer('drawer')"
\end{lstlisting}

\begin{lstlisting}[language=json]
Controller:
{
    "Thought": "The tea caddy has been moved to the edge of the table successfully. Now, I need to close the drawer as per the user's request.",
    "Action": "execute_task()",
    "Action input": {"Task name": "Close the drawer", "Objects": ["drawer"], "Task description": "The robot arm should close the drawer."}
}
\end{lstlisting}

\begin{lstlisting}[language=json]
Executor:
    "close_drawer('drawer')"
\end{lstlisting}

\begin{lstlisting}[language=json]
Controller:
{
    "Thought": "I've moved tea caddy to edge of table and close drawer. The task should be finished.",
    "Action": "finish()",
    "Action input": {"message": "I've moved tea caddy to edge of table and close drawer."}
}
\end{lstlisting}

\end{tcolorbox}

\subsection{Cleaning the table}  \label{appendix real-world clean table}
\begin{tcolorbox}[breakable, colback=Emerald!10,colframe=cyan!40!black,title=\textbf{Conversation log of cleaning the table task}]
\begin{lstlisting}[language=json]
User:
Please clean the table: put rubbish into rubbish bin and put useful objects on the shelf.
\end{lstlisting}
\begin{lstlisting}[language=json]
Controller:
{
    "Thought": "First, I need to analyze the environment to observe the objects on the table, and identify which ones are rubbish and which ones are not.",
    "Action": "observe()",
    "Action input": {"query": "What objects are on the table?"}
}
\end{lstlisting}
\begin{lstlisting}[language=json]
Observer:
{
    "answer": "The objects on the table include a green basket, a remote control, a water bottle, a white paper, and a yellow box.",
    "observed_objs": ["basket", "remote control", "water bottle", "paper", "box"]
}
\end{lstlisting}
\begin{lstlisting}[language=json]
Controller:
{
    "Thought": "Next, I need to identify which objects on the table are rubbish. Usually, papers could be considered as rubbish. For the other objects, I need to confirm.",
    "Action": "observe()",
    "Action input": {"query": "Which objects on the table are rubbish?"}
}
\end{lstlisting}
\begin{lstlisting}[language=json]
Observer:
{
    "answer": "The plastic bottle is rubbish.",
    "observed_objs": ["plastic bottle", "paper"]
}
\end{lstlisting}
\begin{lstlisting}[language=json]
Controller:
{
    "Thought": "Now that I know which objects are rubbish (the plastic bottle and the paper), I need to find the rubbish bin to put them in.",
    "Action": "observe()",
    "Action input": {"query": "Where is the rubbish bin?"}
}
\end{lstlisting}
\begin{lstlisting}[language=json]
Observer:
{
    "answer": "The rubbish bin is in the middle of the table, slightly towards the left side.",
    "observed_objs": ["water bottle", "remote", "paper", "box"]
}
\end{lstlisting}
\begin{lstlisting}[language=json]
Controller:
{
    "Thought": "Now I know where the rubbish bin is. I will start by moving the rubbish into the bin. First, I will move the plastic bottle, followed by paper.",
    "Action": "execute_task()",
    "Action input": {
        "Task name": "Move rubbish into bin", 
        "Objects": ["plastic bottle", "paper", "rubbish bin"], 
        "Task description": "Pick up the plastic bottle and paper, and put them into the rubbish bin"
    }
}
\end{lstlisting}
\begin{lstlisting}[language=json]
Executor:
# Move Bottle to Bin
pick_and_place_object('bottle', get_obj_position('bottle'), get_obj_position('bin'))
# Move Paper to Bin
pick_and_place_object('paper', get_obj_position('paper'), get_obj_position('bin'))}
\end{lstlisting}
\begin{lstlisting}[language=json]
Controller:
{
    "Thought": "I've put the rubbish into the bin successfully. Now, I need to move the remaining objects to the shelf. But first, I need to locate the shelf.",
    "Action": "observe()",
    "Action input": {"query": "Where is the shelf?"}
}
\end{lstlisting}
\begin{lstlisting}[language=json]
Observer:
{
    "answer": "The shelf is on the right side of the table.",
    "observed_objs": ["cup", "remote", "water bottle", "paper", "box"]
}
\end{lstlisting}
\begin{lstlisting}[language=json]
Controller:
{
    "Thought": "Now I know where the shelf is. I will start by moving the remaining objects to the shelf. First, I will move the remote, followed by the box.",
    "Action": "execute_task()",
    "Action input": {
        "Task name": "Move objects to shelf", 
        "Objects": ["remote control", "box", "shelf"], 
        "Task description": "Pick up the remote control and the box, and put them onto the shelf"
    }
}
\end{lstlisting}
\begin{lstlisting}[language=json]
Executor:
# Move the remote control to shelf_1
pick_and_place_object('remote control', get_obj_position('remote control'), get_obj_position('shelf_1'))
# Move the box to shelf_2
pick_and_place_object('box', get_obj_position('box'), get_obj_position('shelf_2'))
\end{lstlisting}
\begin{lstlisting}[language=json]
COntroller:
{
    "Thought": "I've moved all the objects to the shelf and put the rubbish into the bin. The table should now be clean.",
    "Action": "finish()",
    "Action input": {"message": "I have cleaned the table by putting rubbish into the rubbish bin and moving the objects onto the shelf."}
}
\end{lstlisting}

\end{tcolorbox}

\end{document}